\documentclass[12pt]{article}
\date{}

\usepackage{epsfig}
\usepackage{amssymb,amsmath}
\usepackage[mathscr]{eucal}

\numberwithin{equation}{section}

\title{The Parameters of Educability\\
}

\author{Leslie G. Valiant\\
John A. Paulson School of Engineering and Applied Sciences\\
Harvard University\\
Boston, MA 02134\\
{\it valiant@seas.harvard.edu}\\
}
\begin{document}
\date{December 12, 2024}
 \maketitle
\thispagestyle{empty}

\begin{abstract}
The educability model is a computational model that has been recently proposed to describe the cognitive capability that makes humans unique among existing biological species on Earth in being able to create advanced civilizations. Educability is defined as a capability for acquiring and applying knowledge. It is intended both to describe human capabilities and, equally, as an aspirational description of what can be usefully realized by machines. While the intention is to have a mathematically well-defined computational model, in constructing an instance of the model there are a number of decisions to make. We call these decisions {\it parameters}. In a standard computer, two parameters are the memory capacity and clock rate. There is no universally optimal choice for either one, or even for their ratio. Similarly, in a standard machine learning system, two parameters are the learning algorithm and the dataset used for training. Again, there are no universally optimal choices known for either. An educable system has many more parameters than either of these two kinds of system. This short paper discusses some of the main parameters of educable systems, and the broader implications of their existence.

\end{abstract}

\section{Introduction}

The educability model has been recently proposed [V24] to describe humanity's civilization enabling cognitive capability. It aims to characterize this cognitive capability as a mathematically well-defined model. Educability is defined in terms of computational processes that an entity can use to acquire and apply information. It is intended both to describe how humans differ from other present-day species on Earth, and as a description of the potential of machines for emulating human capabilities. 

The notion of {\it intelligence} has never received a widely accepted behavioral definition. We believe that, for that reason, intelligence has not been a particularly useful notion for understanding either humans or machines. On the other hand, the notion of learning has been useful for understanding machines. Learning from examples in order to generalize to new ones can be given a behavioral definition [V84]. Such a definition has served as the criterion for comparing the effectiveness of learning algorithms and thereby also as the strategy for improving them. The practice of AI is now reaping the benefits of learning having a behavioral definition. 

{\it Educability} is a mathematically defined concept having a behavioral definition [V24], in the same sense as learning from examples has, but aims to cover a  broader range of cognitive phenomena. We believe that for that reason it has a chance of being more useful than intelligence has been both for understanding humans and for suggesting new technology. It can be viewed as an attempt to replicate the success of learning from examples to broader phenomena in cognition. 

This paper focuses on the {\it parameters} of the educability model. These are choices that need to be made in implementing the capability of being educable, whether in biology or in a computer. As an example, consider computing. All computers have the same power in the sense of Turing computation, but within that, each actual realization has some parameters, such as the amount of memory available, and the clock rate. As another example, one closer to the current context, in a conventional application of present day machine learning, there are parameters beyond those of the computer on which it is implemented. One is the machine learning algorithm used. Another is the dataset used for training. The parameters mentioned in the case of Turing computation are numerical, and in the case of machine learning, namely the learning algorithm and the dataset, they are non-numerical. Nevertheless, in both cases, the actual choices made for the parameters are critical in their various ways. In the case of machine learning, for example, the choice of learning algorithm determines which datasets will be more easily learned, and the choice of dataset will determine the behavior of the trained classifier.

Educability, as defined, is a more complicated notion than either Turing machines or standard machine learning, and needs many more parameters, both numerical and non-numerical. These parameters reflect the choices that need to be made in designing and educating an educable machine.  For humans some of the basic choices, such as approximate memory capacity, are largely made through evolution. Other choices are made variously in the design of the education system to which the individual has been subject, as well as in the particulars of the educational and life experiences that the individual has had. These parameters are all important, we believe, for understanding both humans and machines. On the one hand, they make explicit some alternative options available in human education,  and, on the other, the potential of machines for emulating human capabilities. 

An example of an important parameter for any standalone cognitive entity that acquires information from its environment is its policy as to what sources of information to trust. We call this its {\it Belief Choice} policy. As psychologists have discovered, humans use a variety of such policies. The best policy appears to depend on the environment in which the individual operates. If there is no universally optimal policy even in just this one parameter, then it already follows that there is no unique metric of a human being “intelligent'' or ``smart''. This illustrates that, for both machines and humans, it is naïve to believe that there is a single metric, such as an IQ test, for evaluating the performance of a human or machine in a general environment.

For brevity, in this paper we assume familiarity with [V24]. Many important questions about educability discussed there are not discussed here. These include: Can meaningful psychological tests for measuring educability in humans be designed? Can human educability be improved by some intervention. What are the lessons for education practice and education science? How did this educability capability come together in the course of human evolution?

The outline of this paper is as follows. In Section 2 we discuss the methodology that we use.  In Section 3 we outline the definition of educability. Sections 4-8 will address five particular parameters that are important for educability. These are Belief Choice, Belief Verification, Teaching to Reason, Management Rules for the Mind’s Eye, and New Concept Formation. Section 9 briefly mentions some additional parameters, and Section 10 is a summary.

A discussion of educability and its parameters offers a perspective from which to understand what machines are doing now and what they can be expected to be capable of in the future. It also offers a perspective from which to understand human behavior.

\section{Scalable and Robust Models of Computation} 

The approach taken here is to regard human cognitive phenomena as capabilities that can be defined as being {\it scalable}, on the one hand, and {\it robust}, on the other. By a {\it scalable computational model} we here mean two things. First, we insist that there is a mathematical specification of what is achieved when the capability is exercised, i.e., there is an input-output specification of what is accomplished. Second, we require that this specification be known to be scalably realizable (in the first instance, realizable by a polynomial time computation.) This ensures that capabilities for which only exponential time (or worse) methods of realization are known are excluded.

The prime example of such a scalable computational model is the universal Turing machine. The task there is, given an arbitrary Turing machine and an input for it, simulate that Turing machine on the universal Turing machine step by step. Both of the above described requirements of a scalable model are satisfied: First, what is to be accomplished in each step of this simulation is mathematically well-defined. Second, each step can be accomplished efficiently, in polynomial time in terms of the relevant parameters, as Turing showed [T36].

As another example, consider the old philosophical problem of induction, discussed by Aristotle and many philosophers since. Here we view this problem from the perspective of a human cognitive capability. This capability involves {\it generalization from experience}. Humans can, through seeing examples of chairs and other objects, acquire the ability to categorize objects they have not seen before as to whether they are examples of chairs. The {\it Probably Approximately Correct} (PAC) model of learning specifies this capability as a computational model [V84]. This formulation also satisfies the two requirements of a scalable computational model stated  above, since (i) there is a mathematical specification of what realizing the capability achieves -- namely scalably accurate generalization to new examples, and (ii) the realization of the capability is defined to require polynomial time. 

It is argued in more detail in [V24] that computational models, whether scalable or not, become really useful if they are, in fact, {\it robust computational models}. Here robustness means that variations in the model that have the same {\it intent} are equivalent. The word variation here is also meant in two different senses. In the first sense, variations in the specification of models that have the same intent should be equivalent in realizing that intent. The main paradigm for this is again the Turing machine. The fact that other variants having the same intent (of defining what can be mechanically computed in the intuitive sense) have been found to have no greater power than standard Turing machines, is the robustness that is clearly crucial in the success of Turing computation in the real world. For the second meaning of variation, one can define (as we also assume for a scalable computational model) the function that is intended to be computed, and then obtain robustness from the many possible ways of computing that function. This second type of robustness is, of course, most meaningful if the specification of what is intended is expressed in different terms from merely describing an algorithm or program, since then it is more meaningful to consider all the different algorithms and programs that realize the intent. The PAC learning model has both kinds of robustness. It has the first kind both as an inheritance from Turing computation, but also more directly from demonstrations that many variants of it have been shown to have the same power [HKLW91]. It gets the second from the fact that there is a specification of what it accomplishes, namely scalably accurate generalization to new examples. 

The general question of why robust computational models can be useful for understanding natural phenomena, is a crucial one. Turing’s 1936 paper [T36] is the singular inspiration here, having brought forth our modern information-based world, containing as it does far more substance far beyond the paper’s explicitly stated mathematical content. The answer to the question is not obvious. Herbert Simon in 1958 predicted that “within ten years most theories in psychology will take the form of computer programs, or of qualitative statements about the characteristics of computer programs” [SN58]. This prediction was not realized, perhaps because one needed to wait for a more nuanced view of what computation has to offer. Such a nuanced view needs to incorporate, for example, the fact that cognition is a large-scale phenomenon that needs scalable mechanisms, ones that are computationally affordable even when the information manipulated is enormous. Both scalability and robustness are at the center of the definition of educability, as they are also at the center of PAC learning.

The recent successes of machine learning in diverse applications we regard as strong empirical positive evidence of the appropriateness of scalable and robust computational approaches to understanding cognition. The study of educability may be viewed as an exploration of the usefulness and power of these approaches to natural phenomena beyond learning from examples.

\section{Outline of Educability}

Educability is defined in [V24] as resting on the following three pillars: (A) learning from experience, (B) being teachable by instruction, and (C) chaining and applying theories obtained in both modes A and B, all three pillars to be realized in a setting that is sufficiently expressive in the following two senses: (i) Information is represented in a {\it Mind’s Eye}, where descriptions of multiple objects and their relationships can be processed, and (ii) flexible use of {\it symbolic names} is supported throughout the system. 

All parts of the above definition of educability have mathematical definitions, but with multiple variants whose differences are specified as parameters. Pillar (A), learning from experience, is formulated as the PAC learning model [V84] and captures the basic process of supervised and self-supervised learning from examples. Present-day AI systems, in essence, realize this one pillar. A currently pervasive application is Large Language Models (LLMs), which can generate smooth and appropriate natural language responses to natural language inputs, after being trained on enormous amounts of natural language text. LLMs are trained in the first instance to predict for a given text the next word or word fragment that is likely to follow. The success of LLMs offers some evidence of the power of learning from examples for realizing behavior that previously had been achieved only by human cognitive activity. 

Pillar (B), the capability of being teachable by instruction, is formulated in terms of universal Turing computation [T36]. It is an ability to absorb an explicitly described formula, recipe, computer program or series of instructions, and, thereafter, to apply it to particular situations.

Pillar (C) is the ability to chain together knowledge acquired at possibly widely different times, whether learned from experience as pillar (A) or explicitly taught as pillar (B), and to apply the chained combination to particular situations. It is formulated in terms of the Robust Logic framework [V00]. Robust Logic incorporates the notion of a Mind’s Eye, which loosely corresponds to the notion of working memory in the human brain. The Mind's Eye is a flexible way of representing a situation that contains multiple objects and some relationships among them. Robust Logic provides a principled way of applying learned knowledge to a situation as represented in the Mind’s Eye and updating that representation so as to represent a consequent situation, perhaps a prediction, or the next situation in the execution of a plan. 

The theory is that a persuasive description of a uniquely human cognitive capability is obtained by combining these three ingredients (A - C) but not by combining any two of them. Any one of the three is a conceptually simple notion, and the three are very different from each other. The richness of the phenomena found in human cognition arises, we believe, from the interplay between such different phenomena. This is analogous to Newton's Laws of mechanics, for example, where the three laws define a much richer system than any one of them alone would.

More detailed definitions of these components can be found in [V24]. This current paper focusses on the issue that educability as just outlined has parameters, and for each parameter a range of options is available. 

\section{Belief Choice}

Educability, as defined, is a powerful capability for acquiring knowledge, information and beliefs from the environment, but nowhere in the definition is there any concomitant capability for verifying whether the knowledge, information and beliefs acquired by instruction, namely by pillar B, are true, valid or useful. This, we believe, constitutes a fundamental vulnerability for any educable entity, including humans, which has to be weighed against the powerful mechanisms for information absorption that educability offers.

The problem is that the entity usually has limited resources and options for checking whether a belief presented is valid, useful, or helpful to the entity. A student in a lecture course on quantum mechanics has limited opportunities to repeat all the experiments on which the theory presented is based. Similarly, a citizen hearing a news story about an event that happened that day on the other side of the world has limited opportunity to rush to that location and verify its truth. 

We believe that this gap for humans is not just an unfortunate omission of human evolution but something that is fundamentally difficult to overcome. Some of the gap is unbridgeable. We might be told, for example, as part of a belief system, that some historical event occurred centuries ago. There are limitations to the extent to which we can achieve certainty about whether it actually happened.

Once a standalone system, whether a biological entity or a machine, can alter its behavior based on information it receives from its environment it needs a policy for determining which external information it should allow to change its behavior and which not. Such a need becomes self-evident if the system can obtain an explicit belief from its environment, as in pillar (B) of our definition of educability, but is already present in PAC learning, pillar (A), since some discretion is needed in deciding which of the examples seen are genuine. Current AI systems are generally not standalone in this sense. The human trainer curates the examples from which the learning algorithm is to train. Similarly, where some knowledge is programmed (such as a knowledge graph in an AI system) a human chooses what knowledge to include. 

We divide the methods an entity might use to address this issue into two broad categories. One is {\it Belief Verification}, which deals with methods that attempt to {\it directly} verify the validity of the belief by analyzing it in the context of further information taken from the world. We shall discuss this in section 5, which follows this section. In the current section we discuss {\it Belief Choice}, which deals with methods that use as criteria the source of the information or a comparison with current beliefs, rather than direct analysis of the substance of the belief.

We focus here on theories that psychologists have offered to explain how humans cope with this ever-present dilemma of Belief Choice. There is a broad variety of such theories. They vary according to whether the choice criterion used is based largely on the source of the information, or whether it is based more on the content. These theories arise from experiments done variously on adults and children.  

Source-dependent belief choice strategies are those where the choice of whether or not to accept a new belief is based on the source of the belief. Current systems of education at all levels are based on trust, with teachers generally deemed worthy of trust for the information they provide. In other settings the situation is less simple, and individuals need to exercise judgement in choosing whether to take someone at their word. Various principles of how these judgements are made have been enumerated. The idea that some individuals are opinion leaders, whose opinions are widely followed, has been advocated [KL55]. Another idea is that we tend to follow the opinions of the social groups we belong to or identify with. In a related direction, there is evidence that when a group takes a position, it may be a riskier one than one that an individual might make. This is known as the group polarization or risky shift theory [S68].

A related phenomenon is trust in authority figures. In the 1960s Stanley Milgram used human subjects who had been told that they were performing word-association tests on some learners [M74]. He instructed the subjects to apply an electric shock to the learner whenever the learner made a mistake. The shocks increased in strength as the experiments progressed and the learners pleaded that the experiment be stopped. The subjects had to decide whether to continue following instructions or to stop. In fact, the learners were Milgram’s confederates, who were not really being subject to electric shocks, but the subjects did not know this. A majority of the subjects carried on, applying what they believed were more and more painful electric shocks.

Content-dependent belief choice strategies are those in which the choice of whether or not to accept a belief depends on the content of the proposed belief. These choice strategies may be dependent also on the beliefs the subject already holds. One example of this is the principle of cognitive dissonance, which posits that humans have a preference for holding sets of beliefs that are consistent with each other rather than contradictory [F57]. In deciding whether to adopt a new belief, when guided by this principle, humans will resist adopting beliefs that are inconsistent with what they already believe. In spite of this tendency, humans do have sophisticated ways of taking new evidence into account when this contradicts what they already believe, as is already apparent in children [G13]. Some related theories are based on cultural background. The cultural cognition of risk theory [KB06] is based on the idea that individuals evaluate the risks of different eventualities according to their cultural beliefs. Hence their evaluation of the desirability of different social changes will vary according to their existing cultural beliefs. There are further theories based on other aspects of the beliefs already held. For example, in answer to the question of why many people reject scientific theories that are widely supported by the expert community, it has been suggested [S17] that some reject such theories simply because the theories are counterintuitive in relation to their previous beliefs.

The above discussion refers to research done on human adults. Children are also standalone entities with the same conundrum as adults of determining whom and what to believe. Much work has been done towards understanding how children cope. In one experiment on children, a child interacts with two adults. There is an object present that is unfamiliar to the child. The two adults give different names to the object. The child has to decide whom to believe. From an early age children generally choose to believe a person they have a strong relationship with, such as a parent, over a stranger. Up to age three, the child would still believe the more familiar person even if that person had previously named objects in a way that the child had known to be wrong. By age five, though, the relative stranger who has a history of being more accurate would be preferred. If there are two strangers to choose between, with similar records of accuracy, then the one closer to the child in culture, for example in accent, would be the one more readily believed [H12].

Clearly any standalone educable machine would need some carefully crafted belief choice policies. Its effectiveness in coping in a complex world could be severely compromised by any failures of this policy.

\section{Belief Verification}

What techniques are available to an entity to verify directly the validity of a belief presented to it?  This question is closely related to the issues discussed in the previous section on Belief Choice. However, in that section, by Belief Choice we meant criteria that referred to the source of the belief or the relationship of its content to what was believed already. The current section is more about techniques that attempt to verify the content of a belief more directly against external realities.

The methods that are available here are, in general terms, along the lines of the stereotypic “scientific method”. When presented with a new belief the entity may, for example, check it against experiences and examples the entity has previously encountered and memorized. The entity may also chain the new belief with beliefs already stored and apply that to a stored example to see whether some contradiction is derived.  The entity may also venture to find further examples in the world, or to perform experiments in the world to provide the new examples that are particularly informative with regard to the belief at hand, much like a scientific experiment.

\section{Teaching to Reason}

Some believe in the unlimited power of reason. Perhaps decisions that turn out to be bad are the result of errors of reasoning. Here we posit that the concept of “reasoning” suffers from similar problems to that of “intelligence” in that no satisfactory definition is known that well characterizes human reasoning. This is a more subtle issue than that for intelligence since for reasoning there is the well-developed field of mathematical logic that is a mathematically rigorous study of it. Mathematical logic has indeed provided a rigorous discussion of mathematical proofs in fixed logical systems and has also proved useful for studying computer programs. However, it has been less successful when applied to human reasoning, both in describing human behavior and also in inspiring computer systems whose behavior is reminiscent of that of humans. Some psychologists and philosophers have sought to characterize the nature of the gap, and hence identify what humans do when they reason that differs from what mathematical logic would recommend [J06, K23].

The view taken here is that there is no fixed reasoning method to be prescribed. Rather, educability provides a generic {\it chaining mechanism} on top of which {\it various kinds} of reasoning mechanisms, or rules of inference, can be implemented. This chaining mechanism, and hence the execution of the rules of inference, are realized in the Mind's Eye. They are realized in terms of updates on representations in the Mind’s Eye. Different kinds of reasoning are represented by different update rules. These update rules may be taught   as pillar B, or learned from examples as pillar A. In addition, some of the rules may be provided at the start, by evolution at birth for a biological entity or by initial design for a machine. 

Examples of reasoning mechanisms that are clearly useful for humans include the following\\
\indent
(i)	Attributing causality to events when that is appropriate. \\
\indent
(ii)	Attributing motives to human individuals and thereby reasoning about their behavior. \\
\indent
(iii) Arguing using counterfactuals, situations that have not occurred.\\
\indent
(iv)	 Performing arithmetic operations on multidigit numbers expressed in positional representation.\\
\indent
(v)	 Assigning probabilities to different eventualities and hence predicting the likelihoods of different outcomes by performing probabilistic calculations. 

As just stated, such update rules may variously be inborn, learned from experience, or learned explicitly by instruction. For example, (ii) above is related to a “theory of mind” and may be widely shared in the animal kingdom. However, some in the above list were discovered and disseminated in the course of human history. An open scientific question is to catalog the reasoning systems that have proved useful for humans and might be useful for machines. We call this section “Learning to Reason" because of the importance we attach to reasoning methods that were discovered by individuals and subsequently widely disseminated via pillar B.

The historical period known as the “Age of Reason” was not the result of an advance in humanity's general reasoning capabilities. It was rather the adoption by a few scientists of specific methods of argument that were effective for deriving a better understanding of physics. 

\section{Management Rules for the Mind’s Eye}

The Mind's Eye is a device in which a situation is represented as a {\it scene} in terms of a number of {\it tokens} and a number of {\it attributes} that apply to subsets of the tokens. An attribute applying to a single token will be a property, such as “green'', or ``dog'', of the entity represented by the token. An attribute that applies to a set of tokens specifies a relationship, such as “inside''. A situation consisting of four entities, with some relationships among them, would be represented as a “scene'' with four otherwise undifferentiated tokens each labelled by the attributes of the entity it represents, and pairs (or larger sets of the tokens) labelled by the relations among them. Given an instance of a situation represented as a scene in the Mind's Eye, the likely consequence of the situation can be simulated by applying  to the representation in the Mind's Eye the various pieces of knowledge or {\it rules} that the entity has acquired via pillar A or B. If a rule is applicable to the scene then the scene will be updated, for example, by adding a new attribute to a token. This update represents a prediction the rule has made for the scene. The Mind's Eye can be viewed as the locus of transient thought.

For such a Mind's Eye to operate, some management rules are needed. For example, there will be a fixed number of tokens in the Mind's Eye, loosely analogous to the “seven plus or minus two" concepts that a human can keep in working memory [M56].  Some heuristic criteria will be needed to determine which tokens to free up from previous uses as the entity progresses from one scene to the next. There is experimental evidence that different primate species have different refresh mechanisms for their working memories, differing in the time scale on which they operate  [A22].

Psychologists and others have speculated, sometimes by introspection, on how we manage in our minds the concepts we are thinking about. For example, Galton [G83] stated: “When I am engaged in thinking anything out... . The ideas that lie at any moment within my full consciousness seem to attract of their own accord the most appropriate out of a number of other ideas that are lying close at hand ... . There seems to be a presence-chamber in my mind .... and an ante-chamber full of more or less allied ideas... . Out of this ante-chamber the ideas most allied to those in the presence-chamber appear to be summoned in a mechanically logical way, and to have their turn of audience." In more recent work on consciousness [B88, BB22], some specifications are given that can be interpreted as heuristic management rules for the Mind's Eye, in the tradition of Galton. In a realization of the educability model, one needs these rules to be precisely defined.

\section{New Concept Formation}

In the construction of an educable system, one needs to start with some base set of attributes that represent concepts. This could be a set of attributes recognized by a human at birth, or a computer when first made. An educable entity will need to be able to enlarge this set through its education. There are various mechanisms through which this enlargement can happen. It is the operation of these mechanisms that we call New Concept Formation. The variety of such mechanisms we can express as parameters. 

Suppose $A_1, …, A_n$ are the current attributes, namely the base attributes as possibly already enlarged by some subsequent education. How can one add more attributes? If there is a teacher available to name a new attribute, say an arbitrary name for $A_{n+1}$, then there is no problem since an educable system can do arbitrary symbolic naming. The entity can then learn from examples labeled with this name or memorize a rule for recognizing the named concept as taught by a teacher. 

However, there are many ways for an educable entity to create new attributes by itself even without a teacher.

One way is to create new attributes that correspond to combinations of two existing attributes, such as $A_3$ AND $A_7$. Which pairs should be chosen? There are many options. One can choose pairs that are somehow surprising, or, in the opposite direction, pairs that occur with higher than expected frequency. The experience of an object or event that has a strikingly surprising combination of attributes may be something we remember. Equally, we may remember some of the statistics of the world, such as that when we are in a certain place the weather is often the same. Whichever of these two selection methods is used, the compound attributes once created can be subsequently treated in the same way as the original base attributes. They can serve both  as features of examples used for learning, and also as the targets of learning. 

We can choose new pairs of attributes in several other ways also. For example, we could create new attributes for all pairs that have occurred, say, at least three, times in past examples. This will enable us to learn a richer set of concepts in the world, but at a greater computational cost.

For constructing compound attributes from existing ones, we may also use quantifiers. Thus $\exists x A(x,y)B(x)$ is a condition on token $y$ that says that there exists another token $x$ with the claimed attributes $A(x,y)$ AND $B(x)$. For instance, we may be defining that token $y$ is an elephant if there is an object $x$ that is a tusk and that is a part of $y$. Here the “there exists" quantifier $\exists$ refers only to single situations as explicitly described together in some scene. Namely, we have to find tokens $x,y$ and the claimed attributes together in scenes represented in the Mind's Eye.

\section{Other Parameters}

The five parameters discussed above are all of fundamental interest to human behavior. Belief Choice and Belief Verification deal with how a standalone system can evaluate whether or not to take outside information seriously. Teaching to Reason deals with how an educable entity can acquire knowledge that is very general and amounts to a general method of reasoning. Management Rules for the Mind's Eye are concerned with higher level heuristics for managing working memory. New Concept Formation offers an avenue to forming new concepts and offers a basic approach to creativity.  These five parameters derive special interest because they may all be influenced by education. As far as humans, it is of interest both to identify the basic inborn mechanisms we have that affect these parameters, but also to investigate what enhancements to performance might be achievable through interventions. Of course, as of yet, we do not know much about which parameters can be modified and what the effect of the various choices for them are.

There are several further parameters beyond these five. The learning process will have a basic set of features or attributes that the entity has methods for recognizing at the outset, such as receptors in the retina in human vision. The learning process will also have a basic learning algorithm and a class of classifiers that will be learned. There may be quantitative limitations on the training time and the number of rules that can be learned. The rules that are taught explicitly will be restricted to some class defined both by the representational language and by quantitative limitations. There will be further parameters that concern the education process itself, most notably the training data for what is learned by example, and the teaching materials that are taught explicitly.

\section{Conclusions}

The notion of intelligence has not served us well in understanding either humans or machines. The fact that no definition has been widely agreed for it is a clue to its weakness. For the same reason, the term artificial general intelligence (AGI) will serve us no better, in this author’s opinion. On the other hand, the concept of learning has served us very well for advancing technology because a behavioral definition (PAC learning) can be given for it that describes a powerful relevant phenomenon. 

The notion of educability is offered to fill some of the gaps that remain in our understanding of cognition. Importantly, it has a behavioral definition. The claim is that it characterizes a powerful phenomenon relevant to cognition. It is formulated to capture the essential cognitive difference between humans and other existing biological species. It intentionally leaves out the many capabilities that we share with other species, such as vision, motor skills and emotions. 

The discussion of parameters, the focus of this paper, highlights what we consider to be the fundamental fact that to construct an educated machine, the same plethora of choices is available as we have in deciding how a person should be educated. We focused in five sections on criteria for what sources to trust, skills to evaluate the veracity of beliefs presented, choice of reasoning methods to teach, management rules for working memory, and skills that help in imagining new possibilities. There are others that are even more obvious. These include the teaching materials used and the teaching methods employed. Of course, when designing a machine, besides making the choices that are made in educating an individual, choices are also available that for humans have been largely determined by evolution.

The multiplicity of parameters is strongly hinted at by the broad variation found in human cognitive activity. Cognitive performance of an educated entity cannot be measured along a one-dimensional axis because of the innumerable parameters or choices made in the education process. Even if all the parameters are fixed except for the teaching materials provided during the education, variation in the latter will lead to educated entities whose performance cannot be meaningfully compared by any one metric.

The multiplicity of parameters is not a bug in the educability model but an integral feature of human cognition.

\end{document}